\documentclass[10pt,twocolumn,letterpaper]{article}
\usepackage[T1]{fontenc}
\usepackage[pagenumbers]{cvpr}
\usepackage{tabularx,array}
\newcolumntype{L}[1]{>{\raggedright\arraybackslash}p{#1}}
\newcolumntype{Y}{>{\raggedright\arraybackslash}X}

\usepackage{cuted} 

\usepackage{currfile} 

\usepackage{caption} 

\usepackage{microtype}

\usepackage{placeins}
\usepackage{flafter}
\usepackage{balance}

\definecolor{cvprblue}{rgb}{0.21,0.49,0.74}
\usepackage[breaklinks,colorlinks,allcolors=cvprblue]{hyperref}

\def\paperID{*****}
\def\confName{CVPR}
\def\confYear{2026}
\newcommand{\methodname}{\textbf{SemantiScene-Edit}}
\title{\methodname{}: Expert-Synthesized Supervision and Dual-Path Source Conditioning for Instruction-Based Video Editing}

\author{\normalsize
\textbf{Xinyu Chen}\textsuperscript{1,2,3,*},
\textbf{Song Wu}\textsuperscript{3,*, \textdaggerdbl},
\textbf{Dingbao Shao}\textsuperscript{1,2,3,*},
\textbf{Jiahang Li}\textsuperscript{1,2,*},
\textbf{Jiang Lin}\textsuperscript{1,2} \\
\textbf{Yuhang Liu}\textsuperscript{1,2},
\textbf{Duo Li}\textsuperscript{1,2},
\textbf{Jiaxin Hu}\textsuperscript{1,2},
\textbf{Shengrong Gu}\textsuperscript{1,2},
\textbf{Liang Li}\textsuperscript{3}, \\
\textbf{Qian Wang}\textsuperscript{3},
\textbf{Junlan Feng}\textsuperscript{3,\textdagger},
\textbf{Ying Tai}\textsuperscript{1,2},
\textbf{Jian Yang}\textsuperscript{1,2},
\textbf{Zili Yi}\textsuperscript{1,2,\textdagger} \\[0.5ex]
{\normalfont\small
\textsuperscript{1}State Key Laboratory of Novel Software Technology,
Nanjing University, Nanjing, China} \\
{\normalfont\small
\textsuperscript{2}School of Intelligence Science and Technology,
Nanjing University, Suzhou, China} \\
{\normalfont\small
\textsuperscript{3}China Mobile Jiutian Artificial Intelligence
Technology (Beijing) Co., Ltd.} \\[0.5ex]
{\normalfont\small
\textsuperscript{*}Equal contribution.\quad
\textsuperscript{\textdaggerdbl}Project leader.\quad
\textsuperscript{\textdagger}Corresponding author.
}}

\begin{document}
\maketitle
\begingroup
\renewcommand{\thefootnote}{}
\footnotetext{%
This work was done while Xinyu Chen and Dingbao Shao were interns at
China Mobile Jiutian Artificial Intelligence Technology (Beijing) Co., Ltd.
}
\endgroup
\begin{abstract}
Instruction-guided video editing requires executing a requested change while preserving the source video's appearance and temporal structure. We present \methodname{}, a video editor that couples a multimodal encoder with a pretrained video diffusion transformer. The encoder jointly processes source frames and the editing instruction, and learnable queries convey their semantic context to the generator through a trainable connector and cross-attention. A complementary visual pathway concatenates source-video latents with noisy target latents before patch embedding, supplying dense spatial and temporal information for content preservation. We train the model with flow matching and a three-stage image--video curriculum comprising semantic alignment, joint adaptation, and high-resolution refinement. To support training, we construct over 1.2 million instruction-conditioned video-editing records covering object addition, removal, replacement, and attribute modification. We also construct a 100-case benchmark for evaluating instruction execution, source preservation, temporal consistency, and visual quality. Comparative experiments show that \methodname{} effectively follows editing instructions while retaining source content. At inference, the model requires only a source video and a natural-language instruction.
\end{abstract}

\section{Introduction}
\label{sec:intro}
Instruction-guided video editing aims to modify a source sequence according to a natural-language command while retaining content outside the requested change. Even changing a person's shirt color requires the model to identify the correct subject, modify its appearance, and preserve pose, background, and motion across frames. Effective editing therefore depends on semantic grounding, detailed source context, and temporally coherent generation. Adapting a video generator to this task requires coordinating these forms of semantic and visual information.

Recent video foundation models, including Wan~\cite{wan2025}, HunyuanVideo~\cite{hunyuan2024}, and CogVideoX~\cite{cogvideox2024}, offer strong generative priors for appearance and motion. Diffusion-based editors such as Tune-A-Video~\cite{tuneavideo2022} and TokenFlow~\cite{tokenflow2023} adapt generative priors to preserve source structure during editing. More recent frameworks, including InstructX~\cite{instructx2025}, UniVideo~\cite{univideo2025}, and Kiwi-Edit~\cite{kiwiedit2026}, connect multimodal understanding with diffusion-based generation. These developments motivate combining instruction semantics with detailed source context and training the generator to use both effectively.

We introduce \methodname{}, an instruction-guided video editor that couples a multimodal large language model (MLLM) with a pretrained video diffusion transformer. Its semantic pathway jointly processes sampled source frames and the editing instruction. Learnable queries summarize this context, and a trainable connector projects their features into the generator's conditioning space. Cross-attention then conveys the requested transformation to the generator in the context of the source scene.

A complementary visual pathway uses a frozen video variational autoencoder (VAE) to encode the source sequence into spatiotemporal latents. These latents are concatenated channel-wise with noisy target latents before patch embedding, giving the generator access to source layout, subject appearance, and motion. Semantic queries represent the editing request, while dense latents retain sequence details for content preservation. Together, they guide the transformation within the source scene. At inference, the model takes only a source video and an editing instruction.

We train this architecture with a flow-matching~\cite{flowmatching2022} objective and a three-stage image--video curriculum. During \emph{semantic alignment}, the diffusion transformer is frozen and dense source conditioning is disabled while the queries, connector, and MLLM LoRA~\cite{lora2022} parameters are trained. During \emph{joint adaptation}, source latents are enabled and the semantic pathway and diffusion transformer are optimized together. Image examples supervise appearance changes, while video examples introduce motion and changing visibility. Finally, \emph{high-resolution refinement} uses manually screened videos to improve fine visual detail. The VAE remains frozen throughout.

To support this curriculum, we construct a video-editing dataset containing 400K records for object addition, 400K for object removal, and over 400K for object replacement and attribute modification combined. Starting from unpaired clips, task-specific experts generate editing pairs, which are matched with instructions describing the realized changes. Semantic and video-quality filtering selects the final source--instruction--target triplets.

We further construct a 100-case video-editing benchmark to evaluate instruction execution, source preservation, temporal consistency, and visual quality. Comparative experiments show strong instruction-following and source-preservation performance, with further room to improve aesthetic and imaging quality.

The main contributions of this report are:
\begin{itemize}
  \item An instruction-guided video editor that combines multimodal semantic queries with dense source-video conditioning and is trained through semantic alignment, joint adaptation, and high-resolution refinement.
  \item A video-editing dataset containing 400K addition records, 400K removal records, and over 400K replacement and attribute-modification records combined, constructed with task-specific experts and quality filtering.
  \item A 100-case video-editing benchmark constructed for this study, covering instruction following, source preservation, temporal consistency, and visual quality.
\end{itemize}
The remainder of the report reviews related work, presents the supporting data construction, details the model architecture and training configuration, and analyzes the experimental results.

\section{Related Work}
\label{sec:related}
\subsection{Diffusion-Based Video Editing}
Video editing has evolved from adapting image diffusion models to developing video-specific methods. Tune-A-Video~\cite{tuneavideo2022} adapts a pretrained image model to a single video using spatiotemporal attention and one-shot tuning. Video-P2P~\cite{videop2p2023} combines inversion with decoupled guidance and cross-attention control. FateZero~\cite{fatezero2023} retains intermediate attention maps during inversion and fuses them during editing to preserve structural and motion information. Together, these methods show how source-dependent representations enable controlled edits beyond text-to-video synthesis.

Temporal propagation offers another route to consistency. TokenFlow~\cite{tokenflow2023} enforces consistency in diffusion feature space using correspondences between frames. Rerender A Video~\cite{rerender2023} translates key frames and propagates their appearance through temporal patch matching and blending. MagicEdit~\cite{magicedit2023} disentangles content, structure, and motion signals, while MotionEditor~\cite{motioneditor2023} uses a content-aware motion adapter and attention injection for motion modification. AnyV2V~\cite{anyv2v2024} separates first-frame image editing from image-to-video generation with temporal feature injection. This decomposition relates to our offline replacement and attribute synthesis, for which \methodname{} trains one editor on the resulting pairs.

End-to-end instruction editors learn the transformation from a source video and a command. InstructVid2Vid~\cite{instructvid2vid2023} constructs video-instruction triplets and introduces an inter-frame consistency loss. VACE~\cite{vace2025} instead formalizes reference, editing, and masking inputs through a Video Condition Unit and a Context Adapter. These systems illustrate different conditioning interfaces. \methodname{} uses masks and edited key frames during data construction, while inference requires only a source video and instruction.

\subsection{Synthetic Paired Supervision}
Paired editing videos are scarce in ordinary collections. Synthetic pipelines address this gap by combining image editors, video generators, and quality filters. Se\~norita-2M~\cite{senorita2025} constructs approximately two million editing pairs using specialized video editors and a filtering pipeline. InsViE-1M~\cite{insvie2025} generates alternative edits of the first frame, screens them with a multimodal evaluator, propagates the accepted edit, and performs further frame-quality and motion checks. Together, these works underscore the need to generate and select training targets.

Ditto~\cite{ditto2025} combines an image editor with an in-context video generator, uses an efficient distilled architecture and temporal enhancer for production, and employs an agent to create instructions and filter outputs. Its Ditto-1M~\cite{ditto2025} dataset trains the Editto~\cite{ditto2025} model with a curriculum. OpenVE-3M~\cite{openve2025} broadens coverage to spatially aligned and non-aligned edits and introduces OpenVE-Bench~\cite{openve2025}. ReCo~\cite{reco2025} combines region-aware latent and attention regularization with a 500K-pair dataset, explicitly modeling the interaction between edited and unedited regions during in-context generation. Together, these works connect dataset scale with semantic coverage, locality, and temporal quality.

Our dataset uses task-specific experts to construct source--instruction--target triplets, followed by semantic and video-quality filtering. Training combines these records with public Ditto~\cite{ditto2025} data, as described in Sec.~\ref{sec:training}.

\subsection{Multimodal Understanding and Unified Editing}
MLLM-conditioned generators use visual context to interpret instructions. InstructX~\cite{instructx2025} studies the integration of MLLMs and diffusion models across image and video editing, including image-to-video transfer and modality-specific features. UniVideo~\cite{univideo2025} combines an MLLM and a multimodal diffusion transformer in a dual-stream architecture for understanding, generation, and editing. Its joint training supports task composition and transfer from image editing. These results motivate using image supervision alongside video pairs, especially when a requested change is largely appearance-based.

Omni-Video~\cite{omnivideo2025} connects an MLLM vision head to a diffusion decoder through a lightweight adapter and staged training. Omni-Video~2~\cite{omnivideo22026} uses explicit target captions and contextual representations to guide generation and compositional editing. Kling-Omni~\cite{klingomni2025} also integrates multimodal inputs within a generalist video creation framework. These systems pass contextual information to the generator through continuous tokens, target descriptions, adapters, or unified multimodal representations, while retaining pretrained visual priors.

Kiwi-Edit~\cite{kiwiedit2026} extends instruction editing with reference guidance. Its RefVIE~\cite{kiwiedit2026} pipeline augments video-editing pairs with synthesized reference images, and its model combines query-based semantic conditioning with source and reference latent features. Our semantic interface follows the query-based MLLM--diffusion design of InstructX~\cite{instructx2025}. We combine this interface with channel-wise source-latent conditioning and a three-stage image--video curriculum for editing from a source video and instruction.

\section{Video Editing Dataset Construction}
\label{sec:data}
\begin{figure*}[t]
  \centering
  \includegraphics[width=\linewidth]{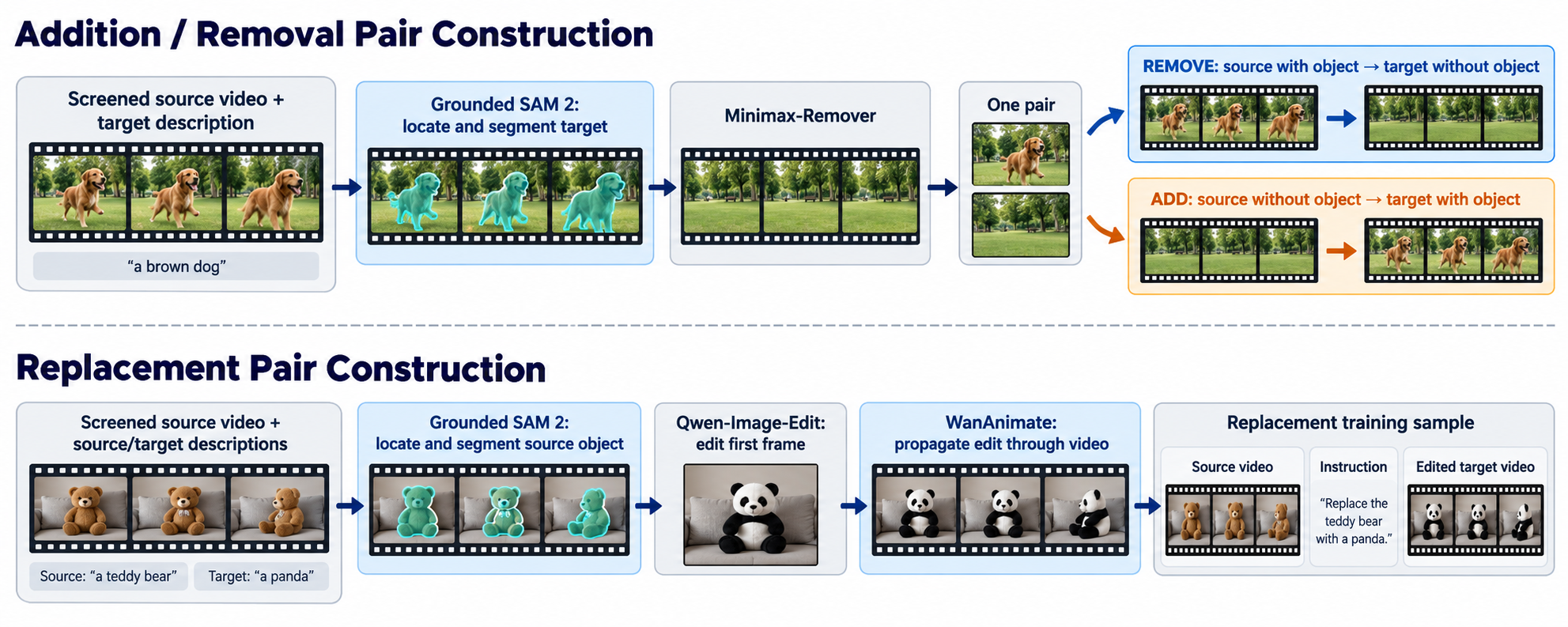}
  \caption{Expert-based paired-video construction. The upper route removes an object and reuses the resulting pair for removal and addition supervision. The lower route localizes the source object, edits the first frame, and propagates the change through the clip for replacement and attribute supervision. Instruction enrichment and triplet-level filtering follow both routes.}
  \label{fig:data}
\end{figure*}

We construct instruction-conditioned video-editing supervision from unpaired source clips. The pipeline screens clips, synthesizes operation-specific source--target pairs, describes the realized edits with instructions, and filters the resulting records. Figure~\ref{fig:data} summarizes the two synthesis routes.

\subsection{Task Representation and Dataset Scale}
A training record is a triplet $R=(S,T,Y)$, where $S$ is the source video, $T$ is the editing instruction, and $Y$ is the edited target. After filtering, the dataset contains 400,000 addition records, 400,000 removal records, and over 400,000 replacement and attribute-modification records combined. Counts refer to instruction-conditioned records; a visual pair can contribute multiple records through directional reuse and instruction variants.

\begin{table}[!htbp]
\centering
\small
\caption{Operation-specific triplet templates. $V^{+}$ contains the selected object, $V^{-}$ is its removed counterpart, and instruction subscripts denote the operation.}
\label{tab:data-routes}
\begin{tabularx}{\columnwidth}{@{}L{0.26\columnwidth}Y@{}}
\toprule
Operation & Triplet template and intended change \\
\midrule
Removal & $(V^{+},T_r,V^{-})$: remove the described object. \\
Addition & $(V^{-},T_a,V^{+})$: introduce the described object in its scene context. \\
Replacement & $(S,T_p,Y)$: replace the source object with a requested target object. \\
Attribute & $(S,T_c,Y)$: modify an appearance property while retaining object identity and scene context. \\
\bottomrule
\end{tabularx}
\end{table}

\subsection{Source Screening and Object Parsing}
Qwen3-VL-235B~\cite{qwen3vl2025} screens candidate clips for visible, identifiable objects, limited occlusion, and scene structure that can be retained. Clips with ambiguous or very small targets, severe occlusion, or excessive complexity are excluded. For each retained clip, the model identifies the object and operation and proposes an initial instruction, including a target identity or attribute when needed.

The initial instruction specifies the intended edit. After synthesis, the instruction is refined using the source and target videos to describe the realized transformation.

\subsection{Operation-Specific Pair Construction}
\paragraph{Removal and addition.}
For removal, Grounding DINO~\cite{groundingdino2023} localizes the described object and SAM 2~\cite{sam22024} produces a mask sequence. MiniMax-Remover~\cite{minimaxremover2025} uses the source video and masks to create an object-free counterpart $V^{-}$ from an object-present clip $V^{+}$. The pair yields two directional records:
\begin{equation}
  \begin{aligned}
  R_r &= (V^{+},T_r,V^{-}),\\
  R_a &= (V^{-},T_a,V^{+}).
  \end{aligned}
  \label{eq:directional-pairs}
\end{equation}
Each direction receives its own instruction and is filtered separately. The checks verify object removal and the spatial and temporal coherence of the reconstructed region.

\paragraph{Replacement and attribute modification.}
For the second route, Grounding DINO~\cite{groundingdino2023} and SAM 2~\cite{sam22024} localize and segment the source object. Qwen-Image-Edit~\cite{qwenimage2025,qwenimageedit2025} applies the replacement or attribute change to the first frame, and Wan-Animate~\cite{wananimate2025} propagates the appearance through the clip to form target $Y$. Sequence-level filtering checks the requested change, preservation of surrounding content, object identity, boundaries, and temporal coherence.

\subsection{Instruction Filtering and Training Integration}
After pair construction, the vision-language model generates instruction variants conditioned on the source, target, operation, and parsed object or attribute. Variants may differ in wording and detail but must describe the same realized transformation. Each complete triplet is then checked for semantic agreement, preservation outside the requested edit, and video quality, including artifacts, blur, boundary defects, flicker, and structural drift. Records are retained only when both semantic and video-quality checks pass.

The accepted triplets provide supervision for all four editing operations. Stage-specific training pools combine this dataset with public editing resources, as described in Sec.~\ref{sec:training}.

\section{Model Architecture and Objective}
\label{sec:method}
\subsection{Conditional Editing Formulation}
Let $S\in\mathbb{R}^{F\times H\times W\times 3}$ denote a source sequence and $T$ its editing instruction. Given training triplets $(S,T,Y)$, the editor learns $p_{\theta}(Y\mid S,T)$ and generates an edited sequence $\hat{Y}$ at inference. The objective is to execute the requested change while preserving source content and maintaining temporal coherence.

\begin{figure*}[t]
  \centering
\includegraphics[width=.96\linewidth]{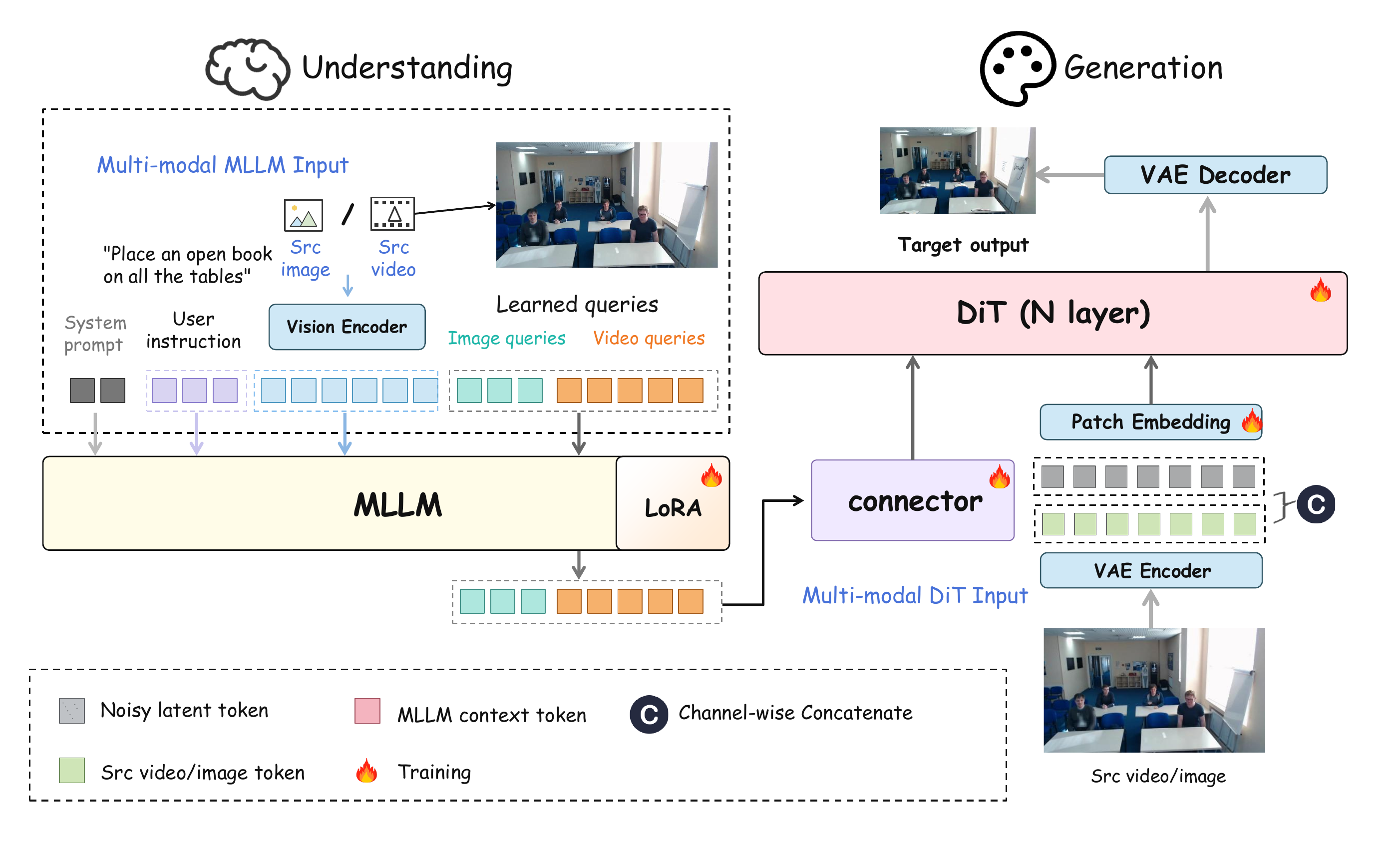}
  \caption{\methodname{} architecture. The multimodal encoder processes source frames, the instruction, and learnable queries to produce semantic features, which a connector projects into the generator's conditioning space. Source-video latents are concatenated channel-wise with noisy target latents before patch embedding, and semantic conditions enter the diffusion transformer through cross-attention. The VAE decoder reconstructs the edited sequence. Flame symbols indicate components optimized during training; their freezing schedule is specified in Sec.~\ref{sec:training}.}
  \label{fig:architecture}
\end{figure*}

\methodname{} uses two conditioning pathways to connect the editing request with the source scene. The semantic pathway summarizes the joint source--instruction context, while the visual pathway encodes the source sequence as dense spatiotemporal latents. Both condition the pretrained generator, as shown in Fig.~\ref{fig:architecture}.

\subsection{Semantic Queries and Dense Source Conditioning}
\paragraph{Multimodal semantic interface.}
The semantic interface follows the query-based MLLM--diffusion design of InstructX~\cite{instructx2025}. Qwen3-VL-8B~\cite{qwen3vl2025} jointly processes sampled source frames, the instruction, and a learnable query sequence. A trainable connector projects the query features into the conditioning space of Wan2.1-14B~\cite{wan2025}, the video backbone. Writing $H_q$ for these features and $P_{\phi}$ for the connector gives
\begin{equation}
  c=P_{\phi}(H_q(S,T)).
  \label{eq:semantic-condition}
\end{equation}
Separate query representations are used for image and video inputs. For a sequence of $N_q$ queries, $H_q\in\mathbb{R}^{N_q\times d_m}$ is projected to $c\in\mathbb{R}^{N_q\times d_c}$, where $d_m$ is the MLLM feature dimension and $d_c$ is the generator's conditioning dimension. The query sequence provides a fixed-length representation for each modality, independent of the input instruction length and number of sampled frames.

Low-Rank Adaptation (LoRA)~\cite{lora2022} adapts the multimodal large language model (MLLM) together with the queries and connector. Their output supplies the semantic condition to the cross-attention layers of the Diffusion Transformer (DiT)~\cite{dit2023}, grounding the requested change in the visible source scene.

\paragraph{Dense source representation.}
The generator operates in a learned compressed space, following latent diffusion~\cite{latentdiffusion2022}. A frozen video variational autoencoder (VAE) encoder $E$ maps the source and target to spatiotemporal latents:
\begin{equation}
  z_s=E(S),\qquad z_y=E(Y).
\end{equation}
The source and target occupy the same latent grid. During editing, the source latent is concatenated with the noisy target state along the channel dimension before patch embedding:
\begin{equation}
  h_0=\operatorname{PatchEmbed}([z_{\sigma};z_s]_{\mathrm{ch}}).
  \label{eq:source-injection}
\end{equation}
For $z_s,z_{\sigma}\in\mathbb{R}^{C\times F_l\times H_l\times W_l}$, concatenation supplies $2C$ input channels and preserves the latent grid $(F_l,H_l,W_l)$. Each patch therefore carries both the current target state and the corresponding source context.

The resulting tokens $h_0$ supply source layout, appearance, and motion to the DiT, while cross-attention to $c$ guides the requested transformation. Paired supervision trains the generator to retain source details where appropriate and modify them where required by the instruction. This combines spatially dense source information with compact semantic guidance throughout generation.

\subsection{Flow-Matching Training Objective}
We train the latent generator with Flow Matching~\cite{flowmatching2022}. Let $\epsilon\sim\mathcal{N}(0,I)$ and let $\sigma\in[0,1]$ denote interpolation time. The target state follows the linear path
\begin{equation}
  \begin{aligned}
    z_{\sigma}&=(1-\sigma)z_y+\sigma\epsilon,\\
    v^*&=\epsilon-z_y.
  \end{aligned}
  \label{eq:flow-state}
\end{equation}
Thus $\sigma=0$ corresponds to the clean target and $\sigma=1$ to noise. The DiT predicts the velocity $v_{\theta}(z_{\sigma},\sigma;c,z_s)$ under the semantic and source conditions. The editing loss is
\begin{equation}
  \mathcal{L}_{\mathrm{FM}}=
  \mathbb{E}\!\left[
    \left\|v_{\theta}(z_{\sigma},\sigma;c,z_s)
    -(\epsilon-z_y)\right\|_2^2
  \right],
  \label{eq:flow-loss}
\end{equation}
where the expectation is over training triplets, Gaussian noise, and interpolation times. The objective minimizes the unweighted squared error between the predicted and target velocities.

The target latent defines the clean endpoint, and the source video and instruction specify the conditional transformation. The VAE remains frozen across training stages, providing a shared latent representation. The semantic interface is first aligned with the frozen generator; dense source conditioning and joint optimization are introduced in subsequent stages, as detailed in Sec.~\ref{sec:training}.

\subsection{Inference Procedure}
Inference computes $z_s$ from the source clip and $c$ from sampled source frames and the instruction. Starting from Gaussian noise, the sampler integrates the predicted velocity from $\sigma=1$ toward $\sigma=0$ under both conditions. A first-order update for the convention in Eq.~\ref{eq:flow-state} is
\begin{equation}
  \begin{aligned}
    z_{\sigma_{k-1}}={}&z_{\sigma_k}\\
      &+(\sigma_{k-1}-\sigma_k)\,
        v_{\theta}(z_{\sigma_k},\sigma_k;c,z_s),
  \end{aligned}
\end{equation}
with decreasing interpolation times. We use a multistep flow solver for sampling and decode the final latent as $\hat{Y}=D(z_0)$.

The source latents and semantic features are computed once per editing request and reused across sampling steps. The sampling configuration is described in Sec.~\ref{sec:training}.

\section{Training Strategy and Configuration}
\label{sec:training}
\subsection{Three-Stage Image--Video Curriculum}
\begin{figure*}[t]
  \centering
  \includegraphics[width=\linewidth]{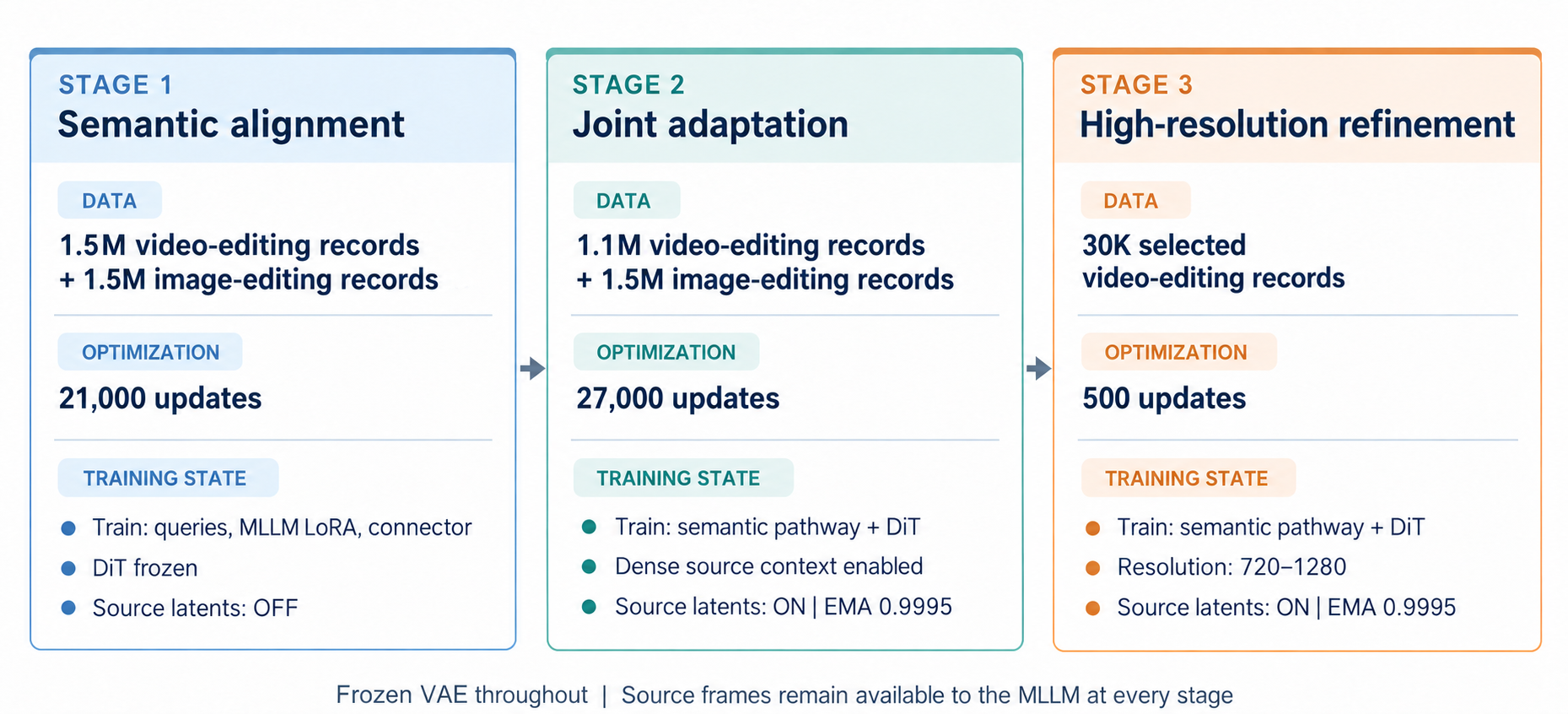}
  \caption{Three-stage curriculum. Semantic alignment updates the query interface with a frozen DiT; joint adaptation enables dense source conditioning and trains the generator; refinement uses a smaller, manually screened collection at higher resolution. Source frames remain available to the multimodal encoder in every stage.}
  \label{fig:training-stages}
\end{figure*}

The three-stage curriculum aligns the semantic interface, adapts the source-conditioned generator, and refines higher-resolution output. Figure~\ref{fig:training-stages} and Table~\ref{tab:training} summarize the schedule. The VAE remains frozen throughout. All stages use a global batch size of 64 and a learning rate of $10^{-5}$.

\begin{table*}[t]
\centering
\small
\caption{Three-stage training schedule. Pool sizes count image- and video-editing records from constructed and public datasets. Update counts are given separately for each stage.}
\label{tab:training}
\begin{tabularx}{\textwidth}{@{}L{0.16\textwidth}YcYL{0.12\textwidth}@{}}
\toprule
Stage & Training pool & Updates & Trainable modules & Source latents \\
\midrule
1: alignment & $\sim$1.5M video + $\sim$1.5M image records & 21,000 & Queries, MLLM LoRA, connector; DiT frozen & Disabled \\
2: adaptation & $\sim$1.1M video + $\sim$1.5M image records & 27,000 & Semantic interface and full DiT & Enabled \\
3: refinement & 30K manually screened video records & 500 & Semantic interface and DiT & Enabled \\
\bottomrule
\end{tabularx}
\end{table*}

\paragraph{Stage 1: semantic alignment.}
The first stage trains the learnable queries, MLLM LoRA parameters, and connector while freezing the diffusion transformer and disabling dense source-latent conditioning. The multimodal encoder still receives source frames and the instruction, allowing the semantic interface to align with the pretrained generator's condition space before the generator is updated.

This stage runs for 21,000 updates on approximately 1.5 million video and 1.5 million image records, with resolutions in the 480--832 range. Video supervision combines our constructed data with public Ditto~\cite{ditto2025} data. Image supervision includes GPT-Image-Edit-1.5M~\cite{gptimageedit2025} and NHR-Edit~\cite{nhredit2026}. Mixing image and video examples introduces appearance changes alongside edits involving motion and changing visibility.

\paragraph{Stage 2: source-conditioned adaptation.}
The second stage enables source latents before patch embedding and trains the full DiT with the semantic interface. It uses approximately 1.1 million screened video and 1.5 million image records for 27,000 updates, initializing the multimodal interface from the Stage 1 checkpoint at step 21,000.

Additional screening selects the video pool for this stage. Joint optimization trains the semantic interface and generator to use dense source context while executing object and attribute changes. An exponential moving average (EMA) model is maintained with a decay of 0.9995.

\paragraph{Stage 3: higher-resolution refinement.}
The final stage uses 30,000 manually screened video-editing records for 500 updates at resolutions in the 720--1280 range. The semantic interface and DiT continue to train with source conditioning enabled. This stage focuses on finer appearance details and higher-resolution output, retaining an EMA decay of 0.9995.

\subsection{Optimization and Implementation}
Stage 2 uses AdamW~\cite{adamw2019} with a constant learning rate after 100 warm-up updates. Table~\ref{tab:implementation} summarizes the optimizer and input settings. Training uses bfloat16 precision, fully sharded data parallelism, and gradient checkpointing to accommodate the multimodal encoder and video generator.

\begin{table}[t]
\centering
\small
\caption{Stage 2 optimization and input settings.}
\label{tab:implementation}
\begin{tabularx}{\columnwidth}{@{}L{0.46\columnwidth}Y@{}}
\toprule
Setting & Value \\
\midrule
Optimizer & AdamW; $\beta_1=0.9$, $\beta_2=0.999$ \\
Learning rate / warm-up & $10^{-5}$ / 100 updates \\
Weight decay / epsilon & 0.03 / $10^{-10}$ \\
Gradient norm limit & 0.05 \\
Precision & bfloat16 \\
Training seed & 42 \\
Video frame window & Up to 81 frames; stride 1 \\
MLLM frame limit & 4 source frames \\
\bottomrule
\end{tabularx}
\end{table}

Resolution bucketing accommodates different aspect ratios and video-token lengths, using sample-size settings of 832 for images and 480 for videos. The generator processes up to 81 consecutive frames, while the semantic encoder receives up to four sampled source frames. The dense latent pathway carries the full training window, and the semantic pathway provides a compact interpretation of its editing context.

\subsection{Inference Configuration}
The inference pipeline combines Wan2.1-14B~\cite{wan2025} with Qwen3-VL-8B~\cite{qwen3vl2025} and loads the EMA generator from the final refinement checkpoint. Sampling uses 20 Flow-UniPC~\cite{unipc2023} steps, classifier-free guidance of 6.0, a flow shift of 3, bfloat16 precision, and seed 43.

The source clip is resized to a 720-pixel long edge with its aspect ratio preserved and dimensions rounded to multiples of 16. Generation processes up to 49 frames, subject to VAE temporal compression, and retains the source playback rate. Semantic features are extracted from a configurable subset of source frames and reused throughout sampling.

\section{Experimental Results and Analysis}
\label{sec:experiments}
\subsection{Comparison Protocol}
We evaluate \methodname{} on our 100-case instruction-guided video-editing benchmark and compare it with UniVideo~\cite{univideo2025} and Kling O1~\cite{klingomni2025}. All methods receive the same source videos and editing instructions. The evaluation covers instruction execution, source preservation, within-video consistency, and visual quality. Table~\ref{tab:results} presents the quantitative comparison.

\subsection{Evaluation Dimensions}
\begin{figure*}[t]
  \centering
  \includegraphics[width=0.82\linewidth]{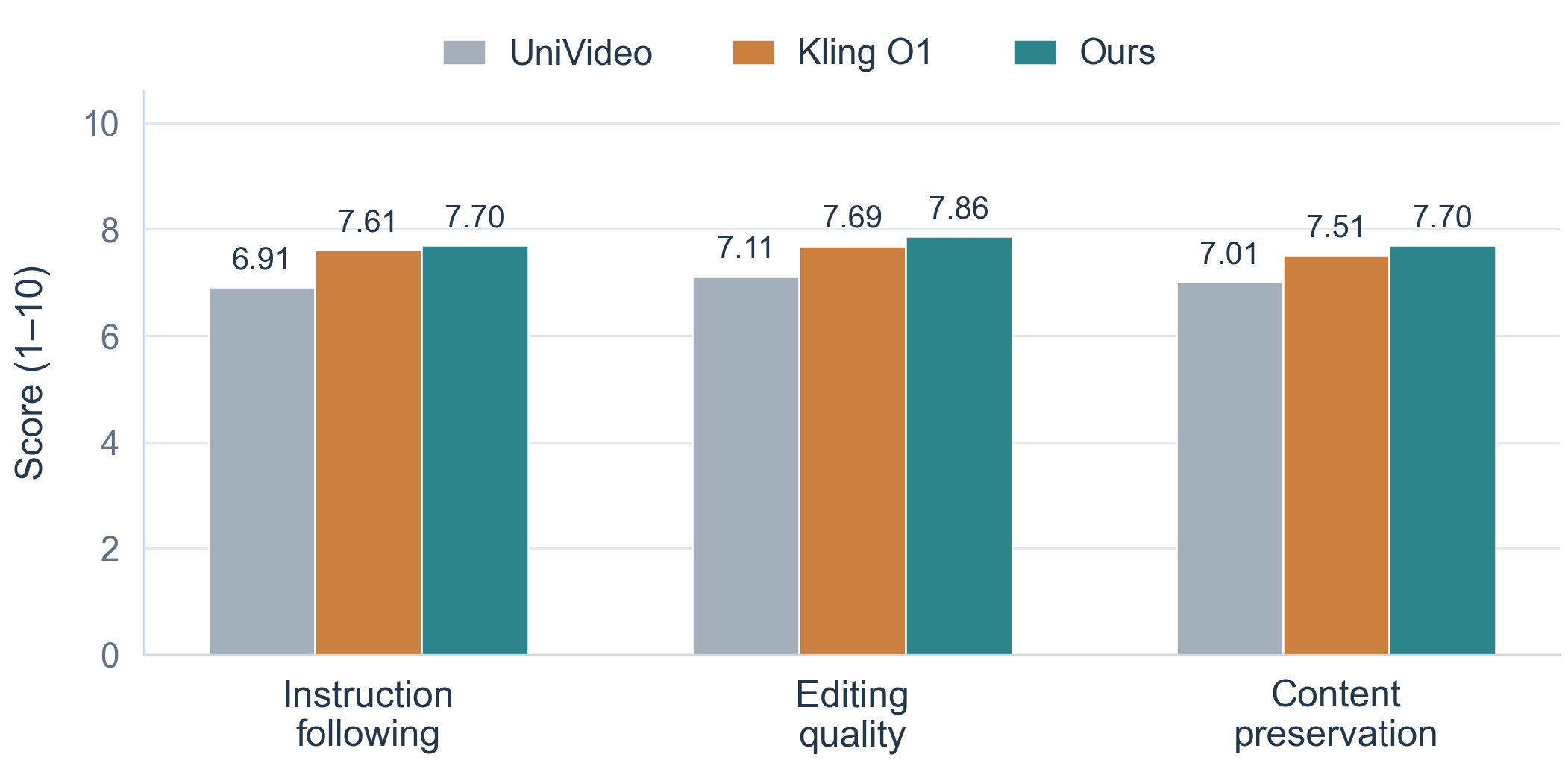}
  \caption{Mean multimodal-evaluator scores for instruction following, editing quality, and content preservation on our benchmark. \textit{Ours} denotes \methodname{}. All three criteria use a 1--10 scale.}
  \label{fig:model-judge}
\end{figure*}

\paragraph{Instruction execution, edit quality, and preservation.}
A Qwen2.5-VL-7B~\cite{qwen25vl2025} evaluator scores instruction following, editing quality, and content preservation from 1 to 10 using the instruction, source video, and generated video. The criteria assess whether the requested transformation is completed, whether it integrates naturally into the scene, and whether surrounding source content is retained. Scores of 1--3, 4--6, and 7--10 correspond to low, partial, and high satisfaction of each criterion. Figure~\ref{fig:model-judge} summarizes the mean scores.

\paragraph{Within-video consistency and visual quality.}
VBench~\cite{vbench2023,vbenchcode} provides background-consistency, aesthetic-quality, and imaging-quality scores. Background consistency measures stability across output frames through normalized CLIP~\cite{clip2021} feature similarities to the preceding and first frames. Aesthetic quality measures visual appeal, while imaging quality evaluates technical properties such as clarity and distortion.

\paragraph{Target-based video metrics.}
For cases with edited target videos, SSIM~\cite{ssim2004} and LPIPS~\cite{lpips2018} measure structural and perceptual agreement, respectively. Generated and target frames are paired by index over their common prefix, with generated frames resized to the target dimensions. Frame scores are averaged within each video and then across evaluated cases.

We also compute VFID-I3D and VFID-ResNeXt as Fr\'echet distances between generated and edited-target video-feature distributions, following the principle of Fr\'echet Video Distance (FVD)~\cite{vfid2018}. Features are extracted with I3D~\cite{i3d2017} and 3D ResNeXt~\cite{resnext2017,resnext3d2018}, respectively. Lower values indicate closer feature distributions.

\paragraph{Supplementary editing score.}
We report an aggregate editing score across three dimensions: instruction compliance, consistency and detail fidelity, and visual quality and stability. Each dimension is scored on a 1--5 scale. The aggregate score is the mean of the three dimensions, averaged across evaluated cases.

\begin{table*}[t]
\centering
\small
\caption{Quantitative comparison on our 100-case benchmark. \textit{Ours} denotes \methodname{}. $\uparrow$ indicates higher is better and $\downarrow$ lower is better; the best value in each row is bold. SSIM, LPIPS, and VFID use cases with edited target videos.}
\label{tab:results}
\begin{tabularx}{\textwidth}{@{}Yrrr@{}}
\toprule
Metric & UniVideo & Kling O1 & Ours \\
\midrule
Instruction following $\uparrow$ & 6.91 & 7.61 & \textbf{7.70} \\
Editing quality $\uparrow$ & 7.11 & 7.69 & \textbf{7.86} \\
Content preservation $\uparrow$ & 7.01 & 7.51 & \textbf{7.70} \\
Background consistency $\uparrow$ & 0.9512 & 0.9455 & \textbf{0.9535} \\
Aesthetic quality $\uparrow$ & 0.5098 & \textbf{0.5248} & 0.5067 \\
Imaging quality $\uparrow$ & 0.6130 & \textbf{0.6973} & 0.6294 \\
Editing score (1--5) $\uparrow$ & 3.797 & 3.942 & \textbf{3.960} \\
VFID-I3D $\downarrow$ & 26.5066 & 36.2155 & \textbf{25.9629} \\
VFID-ResNeXt $\downarrow$ & 1.3501 & 1.0782 & \textbf{0.9253} \\
SSIM (GT subset) $\uparrow$ & 0.5723 & 0.6701 & \textbf{0.7658} \\
LPIPS (GT subset) $\downarrow$ & 0.2399 & 0.2367 & \textbf{0.1966} \\
\bottomrule
\end{tabularx}
\end{table*}

\subsection{Quantitative Analysis}
\paragraph{Task-oriented performance.}
\methodname{} obtains the highest mean scores on instruction following, editing quality, and content preservation, with values of 7.70, 7.86, and 7.70. Relative to UniVideo, the gains are 0.79, 0.75, and 0.69 points; the margins over Kling O1 are smaller. This ordering across the three criteria indicates effective instruction execution with retained source context. The supplementary editing score follows the same ordering, reaching 3.960 for \methodname{}.

\paragraph{Agreement with edited targets.}
On cases with edited target videos, \methodname{} achieves SSIM 0.7658 and LPIPS 0.1966, indicating closer structural and perceptual agreement than either baseline. It also obtains the lowest VFID-I3D and VFID-ResNeXt values, at 25.9629 and 0.9253. Together with the instruction-aware scores, these results show that the generated edits align with the requested transformations and the available targets.

\paragraph{Within-video consistency and appearance.}
\methodname{} achieves background consistency of 0.9535, slightly above UniVideo at 0.9512 and Kling O1 at 0.9455. Kling O1 leads on aesthetic and imaging quality. Compared with UniVideo, \methodname{} has higher imaging quality and slightly lower aesthetic quality. The comparison identifies visual appearance as an area for further refinement alongside the stronger instruction-following and preservation scores.

\begin{figure*}[tp]
  \centering
  \includegraphics[width=\textwidth]{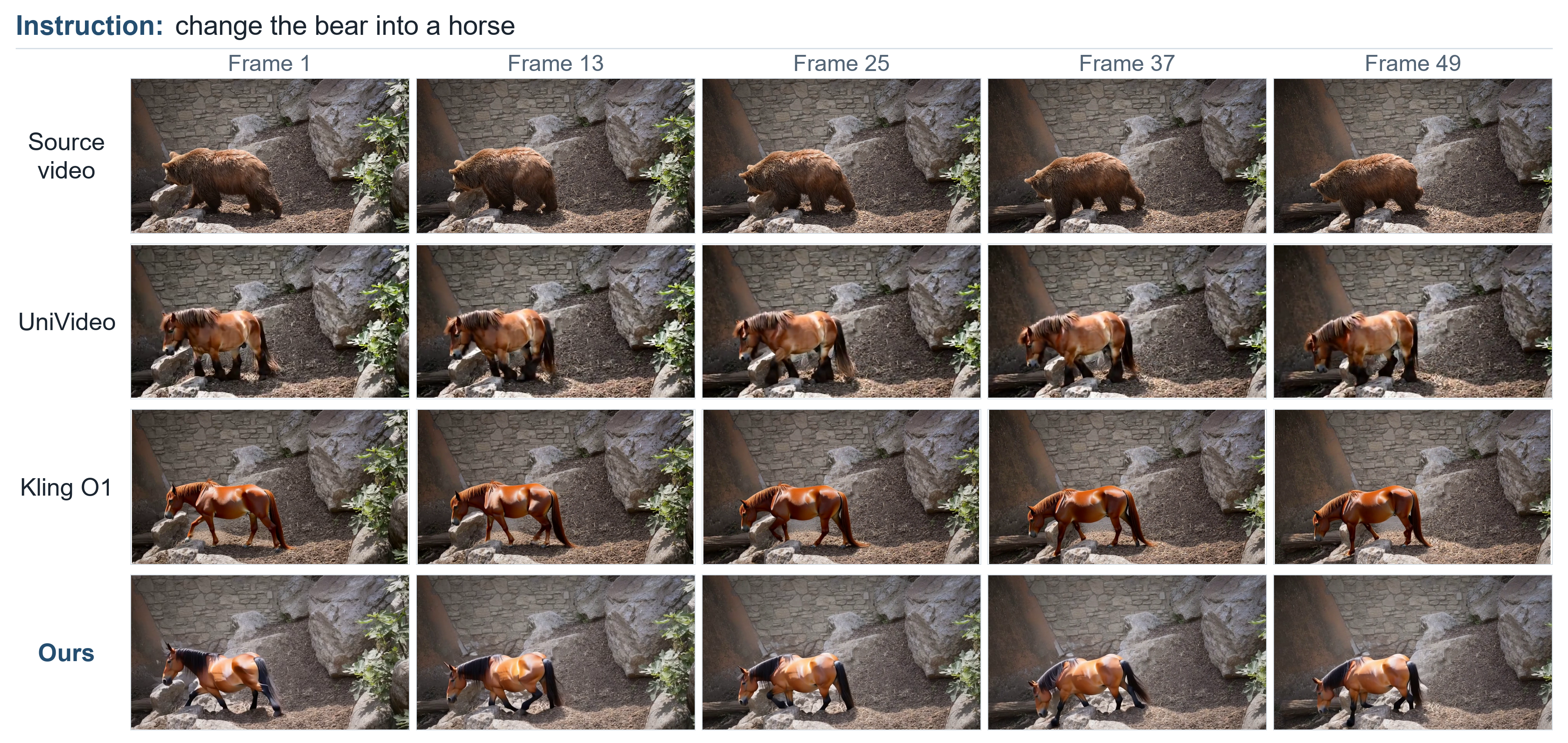}
  \caption{Qualitative comparison for bear-to-horse replacement. Rows show the source video and results from the three editing methods. Columns show frames 1, 13, 25, 37, and 49 from each sequence.}
  \label{fig:qualitative-bear-horse}
\end{figure*}

\begin{figure*}[tp]
  \centering
  \includegraphics[width=\textwidth]{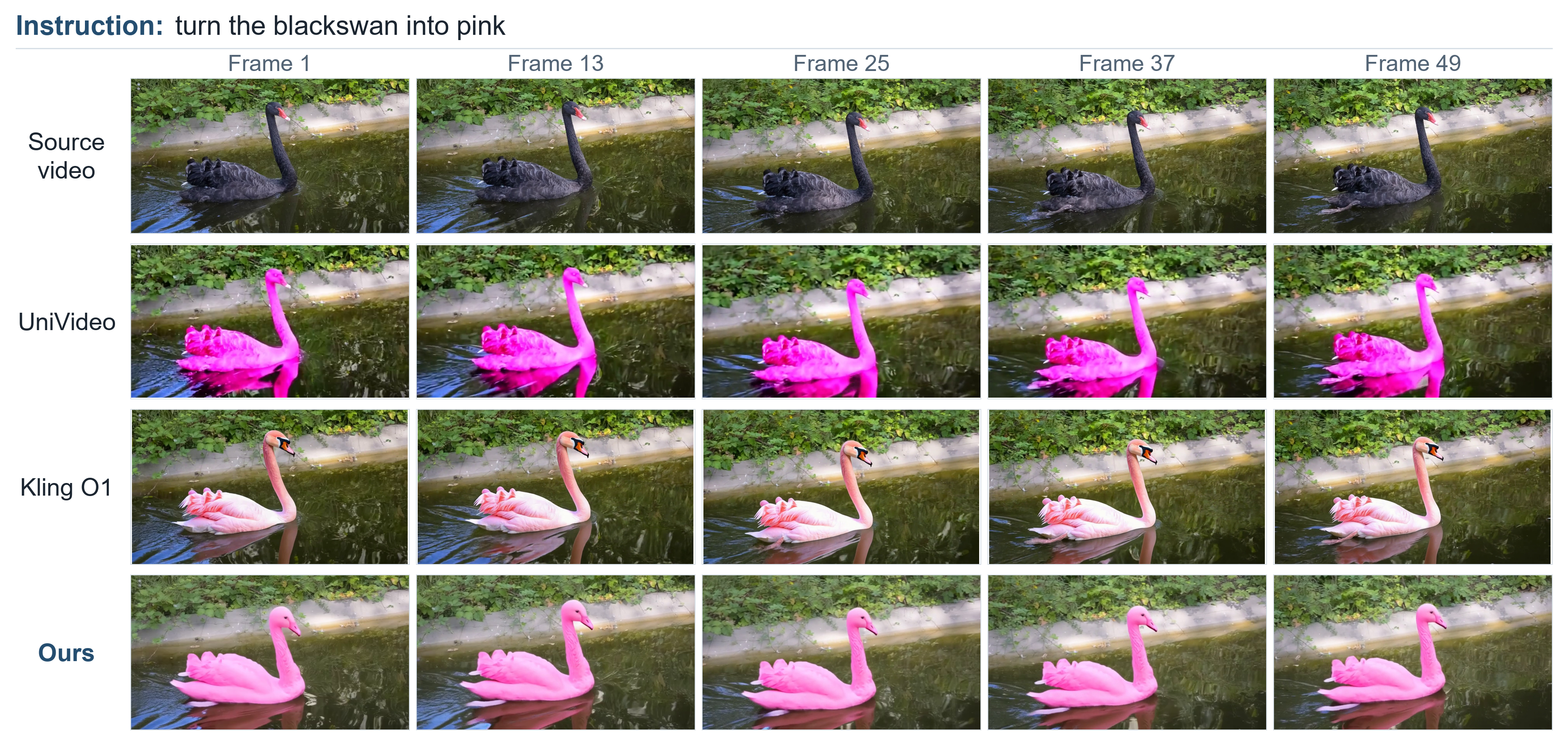}
  \caption{Qualitative comparison for recoloring a black swan to pink. Rows show the source video and outputs from each editing method at frames 1, 13, 25, 37, and 49.}
  \label{fig:qualitative-swan}
\end{figure*}

\begin{figure*}[tp]
  \centering
  \includegraphics[width=\textwidth]{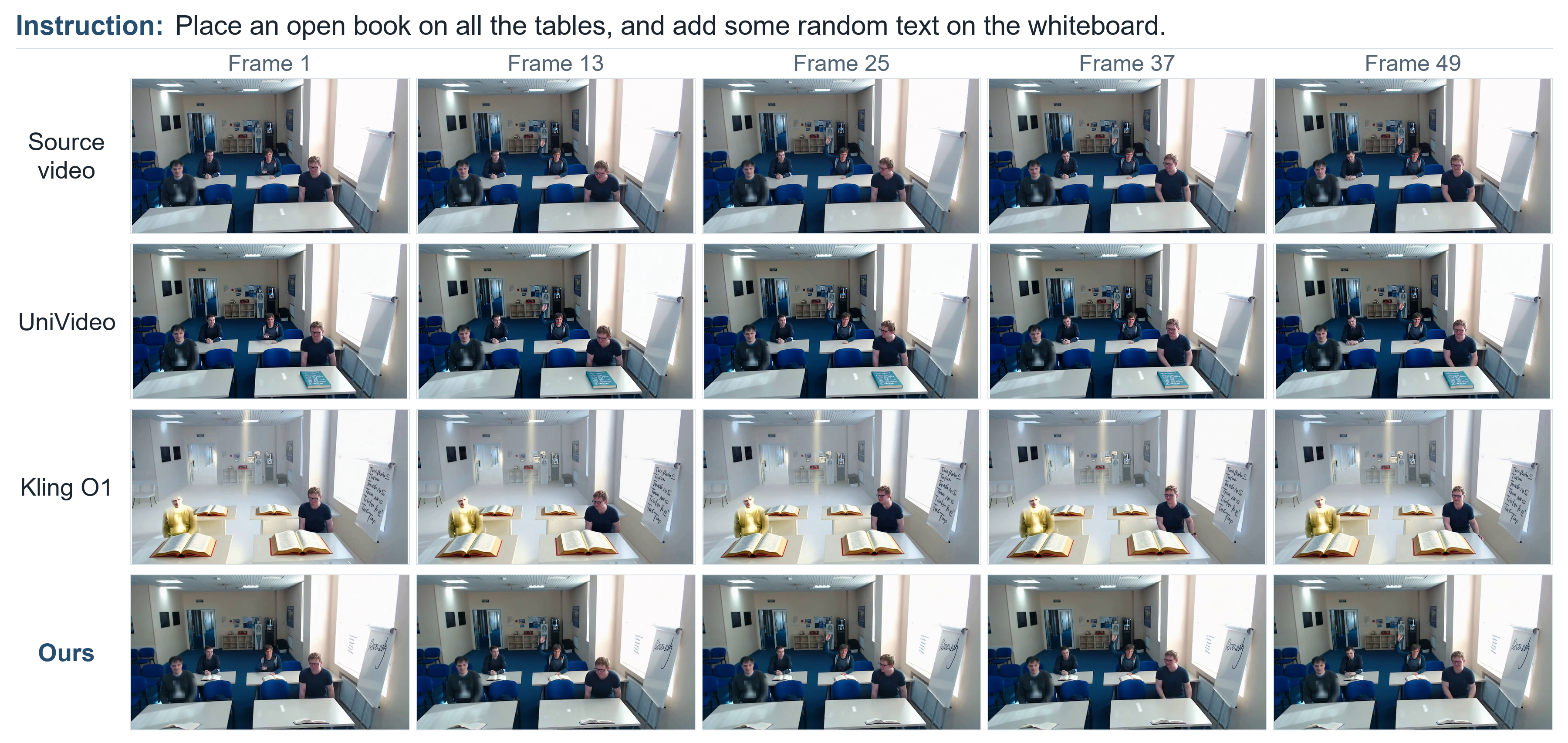}
  \caption{Multi-region addition: placing open books on the tables and adding text to the whiteboard. Rows compare the source video and editing results at frames 1, 13, 25, 37, and 49.}
  \label{fig:qualitative-classroom}
\end{figure*}

\begin{samepage}
\subsection{Qualitative Analysis}
Figures~\ref{fig:qualitative-bear-horse}--\ref{fig:qualitative-classroom} compare three cases covering object replacement, attribute modification, and multi-region addition. Each example presents five sampled frames from the source video and each method's output, together with the editing instruction.
\par
\end{samepage}

\paragraph{Object replacement.}
In the bear-to-horse case (Fig.~\ref{fig:qualitative-bear-horse}), all three methods replace the animal while retaining a recognizable rocky enclosure. \methodname{} preserves the surrounding rocks and the subject's approximate location across the sampled frames, although body shape and pose differ from the source. This example illustrates the geometric changes involved in subject replacement.

\paragraph{Attribute locality.}
In the swan case (Fig.~\ref{fig:qualitative-swan}), \methodname{} produces pink plumage across the sampled frames while retaining the pond and vegetation. UniVideo~\cite{univideo2025} produces a more saturated color, while Kling O1~\cite{klingomni2025} also changes the neck shape and feather appearance. The comparison highlights differences in attribute execution and preservation of the subject's appearance.

\paragraph{Multi-region addition.}
The classroom case (Fig.~\ref{fig:qualitative-classroom}) requests books on the tables and writing on the whiteboard. \methodname{} introduces both elements while largely retaining the people, seating arrangement, and room layout. Kling O1~\cite{klingomni2025} produces prominent open books and writing but also changes the people and interior. This example highlights the balance between executing multiple edits and preserving scene context.

\FloatBarrier
\section{Conclusion}
\label{sec:conclusion}
We presented \methodname{}, an instruction-guided video editor that combines multimodal semantic queries with dense source-video conditioning in a pretrained diffusion transformer. A three-stage image--video curriculum aligns the semantic interface, adapts the source-conditioned generator, and refines higher-resolution output. We also constructed a paired dataset for object addition, removal, replacement, and attribute modification, together with a 100-case benchmark.

Comparative experiments show strong instruction following, editing quality, and source preservation, as well as closer agreement with edited targets. Visual appearance and subject geometry remain areas for further improvement. The framework supports these editing operations through a common source-video and instruction interface.

\clearpage
{\small
  \balance
  \bibliographystyle{ieeenat_fullname}
  \bibliography{main}
}
\end{document}